\pdfoutput=1
\documentclass[11pt]{article}
\usepackage{authblk}

\usepackage{emnlp2021}

\usepackage{times}
\usepackage{latexsym}
\usepackage{textcomp}
\usepackage{amsmath}
\usepackage{graphicx} 
\usepackage{caption,subcaption}
\usepackage{geometry}
\usepackage{tabu}
\usepackage{multirow}
\usepackage{lipsum,multicol}
\usepackage{hhline}

\usepackage[T1]{fontenc}
\usepackage[utf8]{inputenc}

\usepackage{microtype}

\title{Evaluating Predictive Uncertainty under Distributional Shift on\\Dialogue Dataset}

\author{\textbf{Nyoungwoo Lee}}
\author{\textbf{ChaeHun Park}}
\author{\textbf{Ho-Jin Choi}}
\affil{KAIST, Daejeon, South Korea}
\affil{\textit {\{leenw2, ddehun, hojinc\}@kaist.ac.kr}}
\begin{document}
\maketitle
\begin{abstract}
In open-domain dialogues, predictive uncertainties are mainly evaluated in a domain shift setting to cope with out-of-distribution inputs. However, in real-world conversations, there could be more extensive distributional shifted inputs than the out-of-distribution. To evaluate this, we first propose two methods, Unknown Word ($UW$) and Insufficient Context ($IC$), enabling gradual distributional shifts by corruption on the dialogue dataset. We then investigate the effect of distributional shifts on accuracy and calibration. Our experiments show that the performance of existing uncertainty estimation methods consistently degrades with intensifying the shift. The results suggest that the proposed methods could be useful for evaluating the calibration of dialogue systems under distributional shifts.
\end{abstract}

\section{Introduction}
\label{introduction}

Uncertainty estimation in open-domain dialogues has been widely studied~\citep{estimation_tod, feng-etal-2020-none, van-niekerk-etal-2020-knowing, penha2021calibration}, but the predictive uncertainty is primarily evaluated in the domain shift setting~\cite{penha2021calibration}. This setting has an extreme distributional shift to assess the quality of uncertainty of the model under the out-of-distribution inputs. However, real-world conversations might have more extensive distributional shifted inputs than only the out-of-distribution~\cite{NEURIPS2019_8558cb40}. Indeed, safe deployment of models requires robustness in extensive distributional shifts~\cite{amodei2016concrete}. Such capability is also essential for the dialogue systems participating in human conversations.

There has not previously been a rigorous empirical comparison of uncertainty estimation following the gradual distribution shifts of dialogue datasets. Just as making distributional shifts of the image dataset with corruption or perturbation of the source image~\cite{hendrycks2019benchmarking}, the dialogue dataset can also have gradual distributional shifts through the corruption of the source dialogue. For example, causes of uncertainty, such as incompleteness, ambiguity, and inaccuracy, may be inherent in the dialogues due to insufficient context or unknown words.

\begin{figure}[t]
\centering
\begin{tabular}{c}
     \includegraphics[width=0.48\textwidth]{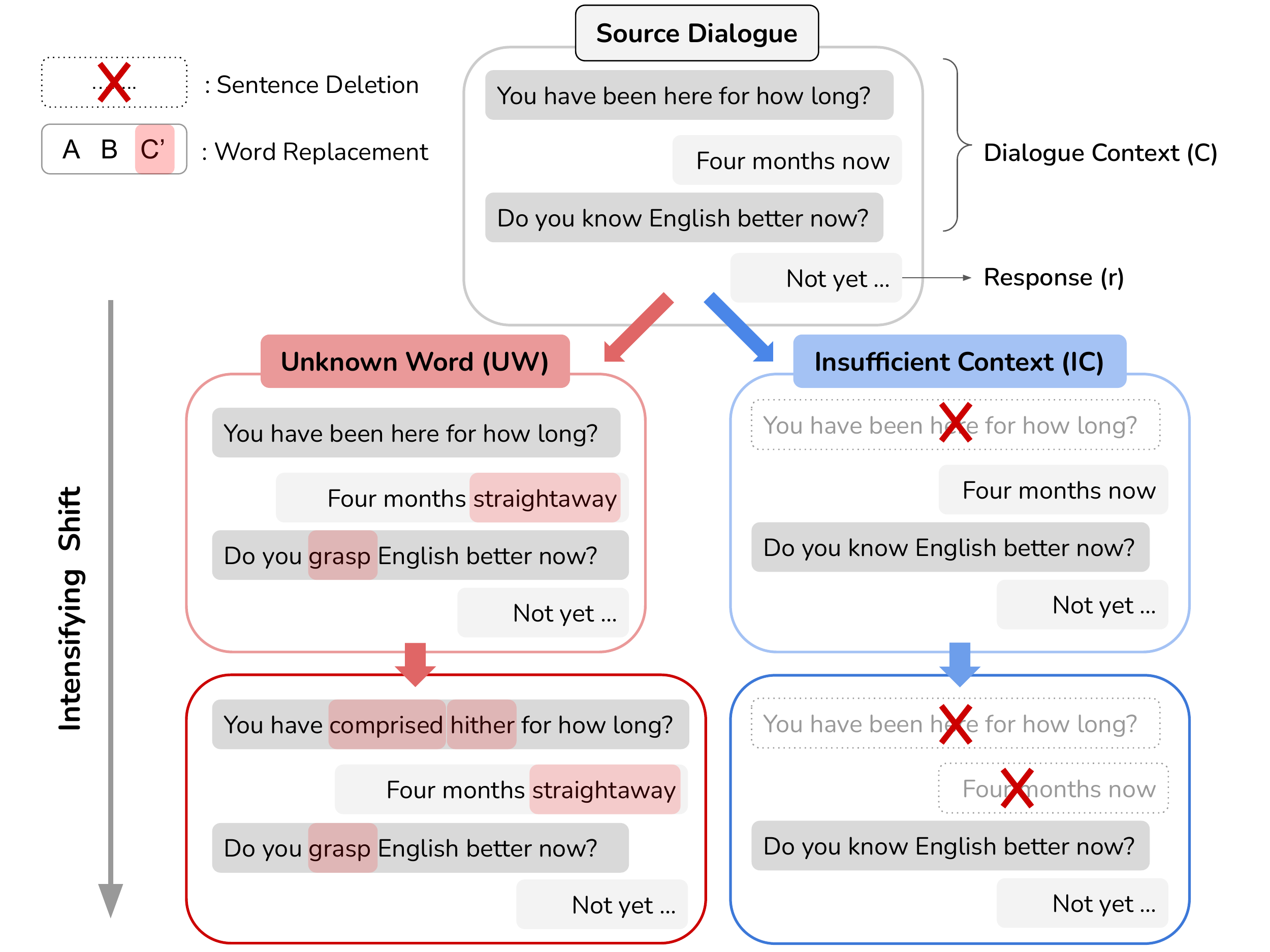}
\end{tabular}
\caption{Examples of distributional shifted dialogue dataset using Unkown Word ($UW$) and Insufficient Context ($IC$).}
\label{fig:method_example}
% \vspace{-0.2in}
\end{figure}

In this paper, we investigate the effect of distributional shifts on the accuracy and calibration of existing uncertainty estimation methods in the dialogue response ranking task. To study this work, we propose two methods, Unknown Word ($UW$) and Insufficient Context ($IC$), to make distributional shifts from the dialogue dataset. Each approach enables gradual changes in the distribution of dialogue data with information corruption in word-level and sentence-level, as shown in Figure~\ref{fig:method_example}.

Experimental results show that the accuracy and calibration of the model downgrade as intensifying the distributional shifts. Such results suggest that well-calibration on the validation and training datasets does not guarantee calibration for distributional shifts. Nonetheless, we find that the ensembles~\cite{lakshminarayanan2016simple} outperforms other methods in terms of accuracy and calibration. Besides, the post-hoc calibration method, temperature scaling~\cite{guo2017calibration} could be an efficient solution to maintain robustness under distributional shift.

\section{Method}
\label{method}

We shift the existing open-domain dialogue dataset using two methods, Unknown Word ($UW$) and Insufficient Context ($IC$), which make word-level and sentence-level changes (see Figure~\ref{fig:method_example}). In this section, we refer to the existing dialogue dataset as the source dataset.

\subsection{UW: Unknown Word}
\label{method_UW}
The first method is to replace target words in the dialogue context with unknown words, making gradual distributional shifts with word-level changes. Just as it is hard to answer clearly when a human faces unknown words in a conversation, the appearance of unknown words can increase the uncertainty. We predict that shifting data in the direction of growing uncertainty will affect the calibration and accuracy of dialogue systems. Since pre-trained language models can build relational knowledge of words during the training phase~\cite{petroni-etal-2019-language}, we assume words that did not appear in the training phase are unknown words.

\paragraph{Word Replacement}
We aim to maintain the meaning of source dialogues as seamless. Thus, we use a synonym dictionary of Wordnet~\cite{wordnet} to replace the target words with synonyms. Since the synonym dictionary may return multiple synonyms, we choose one unknown synonym according to the following rules. If the rule could not find an alternative synonym, we look for another target word.
\begin{enumerate}
    \item Exclude the synonyms of numbers (e.g., 2 $\rightarrow$  two, Two)
    \item Exclude the synonyms consisting of two or more syllables (e.g., computer $\rightarrow$ data-processor, computing-machine) 
    \item Exclude the synonyms that appear in the training dataset
    \item Select the one synonym with the farthest Levenshtein distance among the remaining synonyms~\cite{yujian2007normalized}
\end{enumerate}

\paragraph{Select Target Words to Replace}
We consider the following two conditions when selecting the target words to replace.
\begin{itemize}
    \item \textbf{Replacement Ratio}: The ratio of the target words to the total words in the dialogue context.
    \item \textbf{Word Importance}: The higher it is, the more important word for understanding the dialogue context.
\end{itemize}
To measure the importance of words, we use pre-trained BERT~\cite{devlin-etal-2019-bert}, which fine-tunes as a training dataset of the source dataset. The attention weight of the target word(= token) for the given dialogue context is called the word importance. We shift the source dataset by controlling the two conditions.

\subsection{IC: Insufficient Context}
\label{method_IC}
We then consider sentence-level changes, which deletes sentences in the source dialogue context. We predict that the more sentences are deleted from the dialogue context, the higher the uncertainty of the dialogues, which will affect the calibration and accuracy of the dialogue system.

\paragraph{Sentence Deletion}
We delete the initial sentences except those close to the dialogue's response to keep semantic connections between sentences.

\paragraph{Select Target Sentences to Delete}
We consider the following condition when deleting sentences.
\begin{itemize}
    \item \textbf{Deletion Ratio}: The ratio of the target sentences to the total sentences in the dialogue context.
\end{itemize}
Let's suppose there is a dialogue instance, $D = \{C, r\}$ consisting of $C = \{u_1, u_2, u_3, u_4, u_5\}$ and a response $r$. If the deletion ratio is set to 20\%, we delete the first two sentences from the original context $C$. Then we use the remaining context $C' = \{u_3, u_4, u_5\}$ and the $r$ as the  shifted dialogue instance $D' = \{C', r\}$. We shift the source dataset by controlling the sentence deletion ratio.

\section{Experimental Setup}
\label{experimentalsetup}

\subsection{Dialogue Response Ranking}
We consider the dialogue response ranking task~\cite{wu2017sequential, qu2018analyzing, gu2020utterance, penha2020curriculum}, which is mainly used for evaluating predictive uncertainty of the dialogue systems. This task concerns retrieving the best response for the given dialogue context, by the ranking model training the relationship between dialogue context and its response.

\subsection{Models}
\paragraph{BERT-based Ranker}
We design a BERT-based ranker~\cite{vig2019comparison} for learning the relationship between dialogue context and response. The input for BERT is the concatenation of the context $C$ and the response r, separated by $[SEP]$ tokens. We then make predictions as follows: $f(C, r) = FFN(BERT_{CLS}(concat(C,r)))$.

\paragraph{w/Dropout}
For the dropout~\cite{gal2016dropout}, we use an the BERT-based ranker and employ dropout at the testing time. For predictions, we generate predictions of relevance by conducting $N$(=5) forward passes. At this time, we assume that the mean and variance of N predictions are predictive probability and uncertainty.

\paragraph{w/Ensembles}
For the deep ensembles~\cite{lakshminarayanan2016simple}, we train $M$(=5) BERT-based rankers using different random seeds on the same training dataset. We also assume that the mean and variance of the M predictions are predictive probability and uncertainty.

\paragraph{w/Temperature Scaling}
We also use a post-hoc calibration by temperature scaling~\cite{guo2017calibration} using a validation set.

\subsection{Evaluation Metrics}
\paragraph{Accuracy (Recall@1) $\uparrow$}
To evaluate the retrieval accuracy, we use the recall at 1 out of 10 candidates consisting of 9 candidates randomly chosen from the test set and 1 ground-truth response, called R@1/10.

\paragraph{Brier Score $\downarrow$}
The Brier score is a proper scoring rule for measuring the accuracy of predicted probabilities~\citep{brier1950verification, brocker2009reliability}. It is computed as the squared error between a predicted probability vector and the one-hot encoded true response label. It is mainly used for classification tasks, but the response ranking task can also be assumed to be a classification task that determines whether a given candidate is correct or not.

\paragraph{Empirical Calibration Error (ECE) $\downarrow$}
To evaluate the calibration of neural rankers we use the Empirical Calibration Error (ECE)~\cite{naeini2015obtaining}. It can measure the correspondence between predicted probabilities and accuracy. It is computed by the average gap between within bucket accuracy and within bucket predicted probability for $C$ buckets~\cite{penha2021calibration}.

\subsection{Dataset}
We use DailyDialog~\cite{li2017dailydialog}, which is widely used in open-domain dialogue research. The dataset contains 11,118 training and 1000 validation and test dialogue instances. We conduct a series of experiments under dataset distributional shifts, using the test dialogue instances.

\subsection{Implementation Details}
We fine-tune the pre-trained BERT (\textit{bert-base-cased}) of the \textit{huggingface-transformers}~\cite{wolf2019huggingface} with the DailyDialog training dataset. In the training phase, we configure the training instances so that all turns except the first turn can be a response. We use a cross-entropy loss for 10 candidates consisting of 9 random negative samples and 1 ground-truth response.  We then use the Adam optimizer~\cite{kingma2014adam} with $lr=2e^{-5}$, and we train with a batch size of 12 and fine-tune the model for 1 epoch.

\section{Experiments and Results}
\label{experiments}

\begin{figure*}[t]
    \centering
    \begin{subfigure}[b]{0.32\textwidth}
        \centering
        \includegraphics[width=\textwidth]{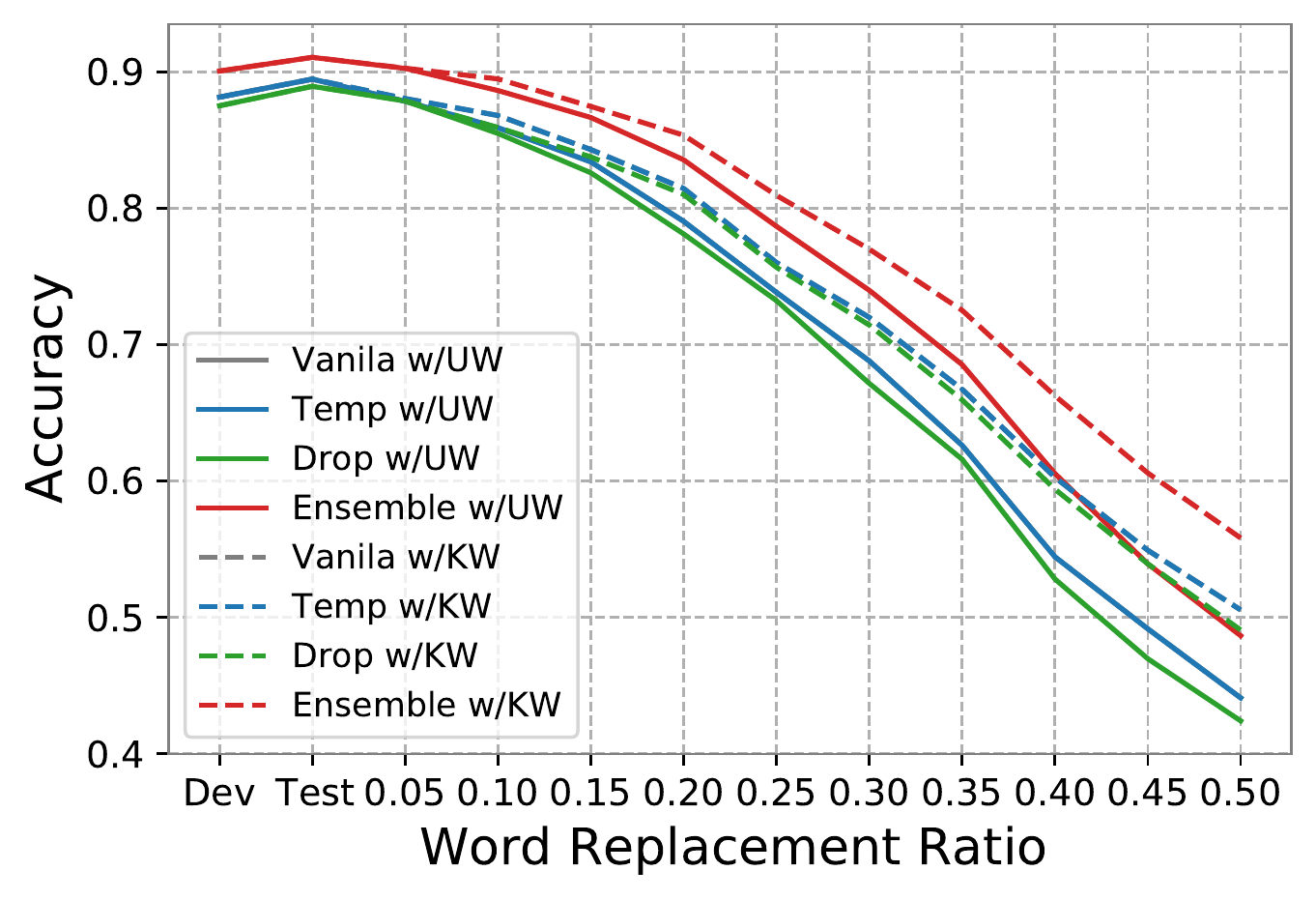}
        \caption[]%
        {{\small Replacement Ratio vs Accuracy}}    
        \label{fig:UW_acc}
    \end{subfigure}
    \hfill
    \begin{subfigure}[b]{0.32\textwidth}  
        \centering 
        \includegraphics[width=\textwidth]{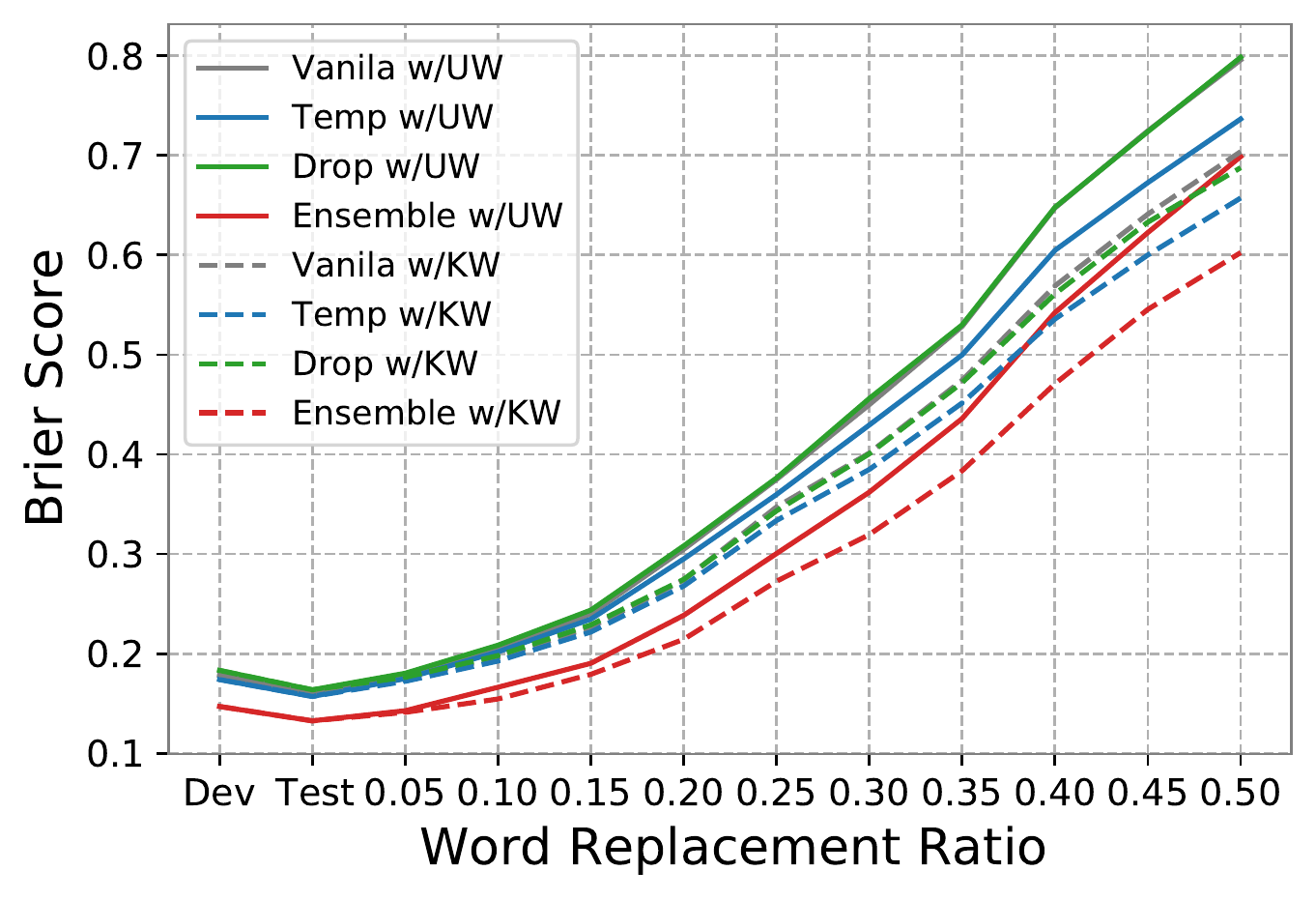}
        \caption[]%
        {{\small Replacement Ratio vs Brier Score}}    
        \label{fig:UW_brier}
    \end{subfigure}
    \begin{subfigure}[b]{0.32\textwidth}
        \centering
        \includegraphics[width=\textwidth]{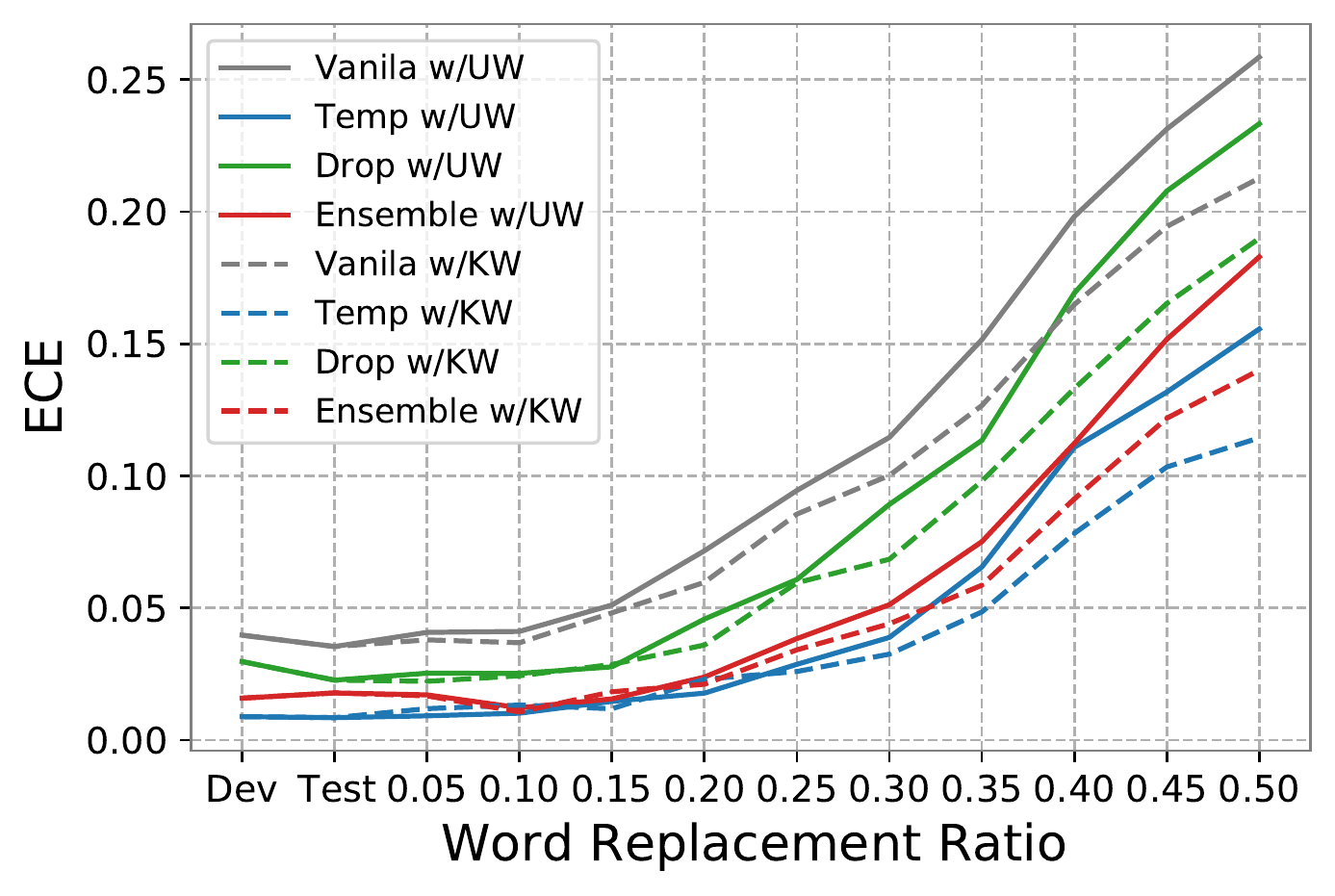}
        \caption[]%
        {{\small Replacement Ratio vs ECE}}    
        \label{fig:UW_ece}
    \end{subfigure}
    % \vskip\baselineskip
    \begin{subfigure}[b]{0.32\textwidth}   
        \centering 
        \includegraphics[width=\textwidth]{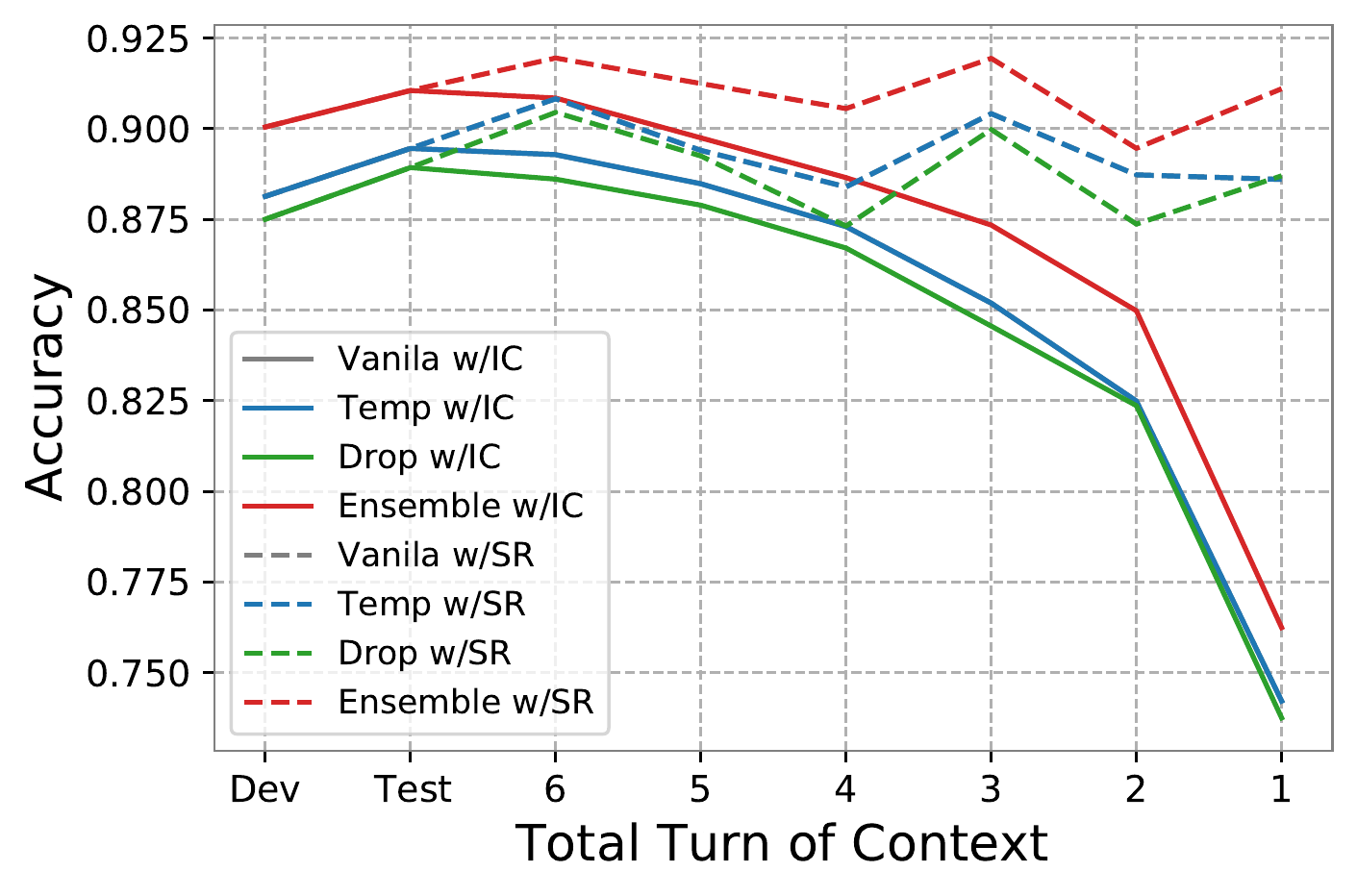}
        \caption[]%
        {{\small Deletion Ratio vs Accuracy}}    
        \label{fig:IC_acc}
    \end{subfigure}
    \hfill
    \begin{subfigure}[b]{0.32\textwidth}   
        \centering 
        \includegraphics[width=\textwidth]{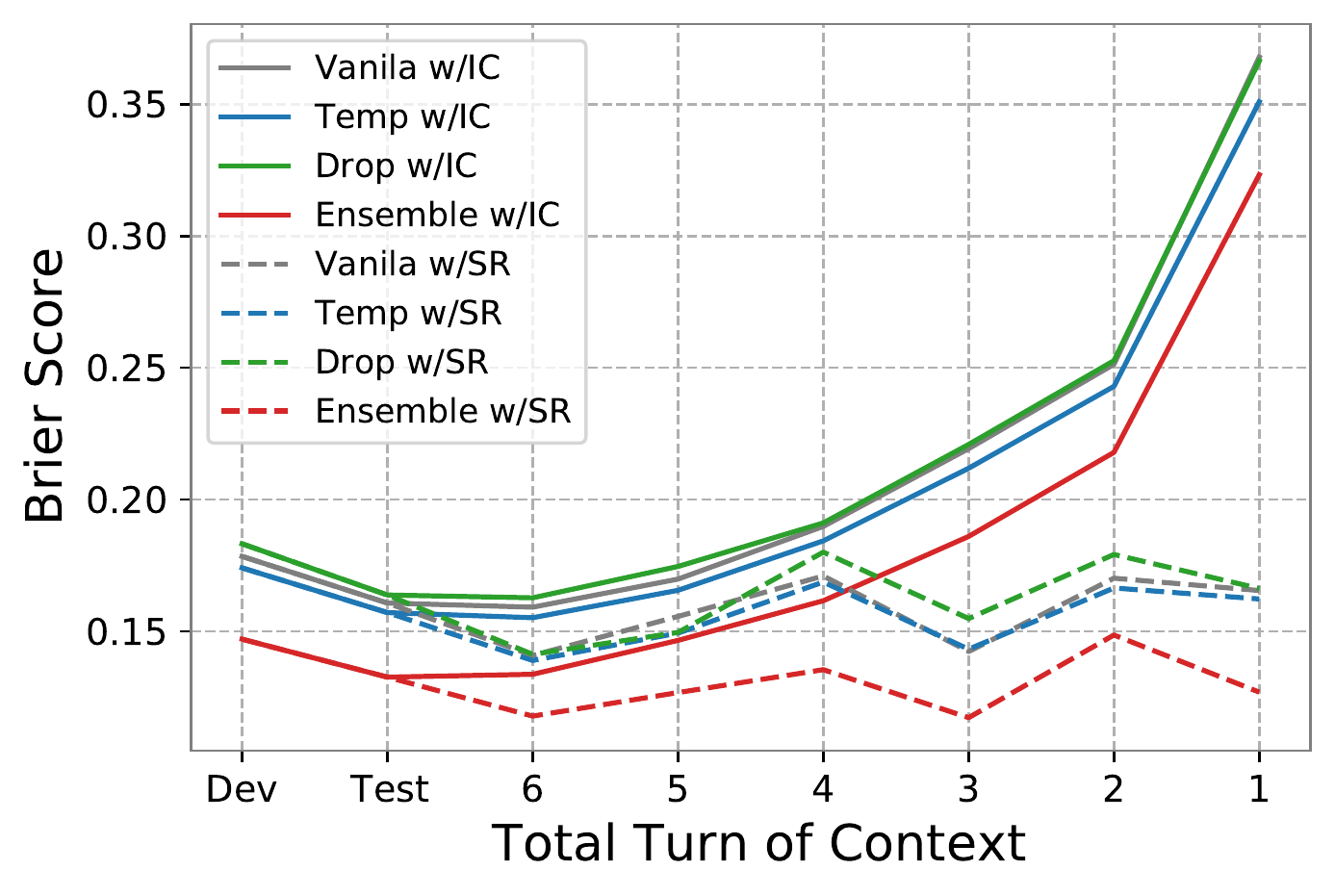}
        \caption[]%
        {{\small Deletion Ratio vs Brier Score}}    
        \label{fig:IC_brier}
    \end{subfigure}
    \begin{subfigure}[b]{0.32\textwidth}   
        \centering 
        \includegraphics[width=\textwidth]{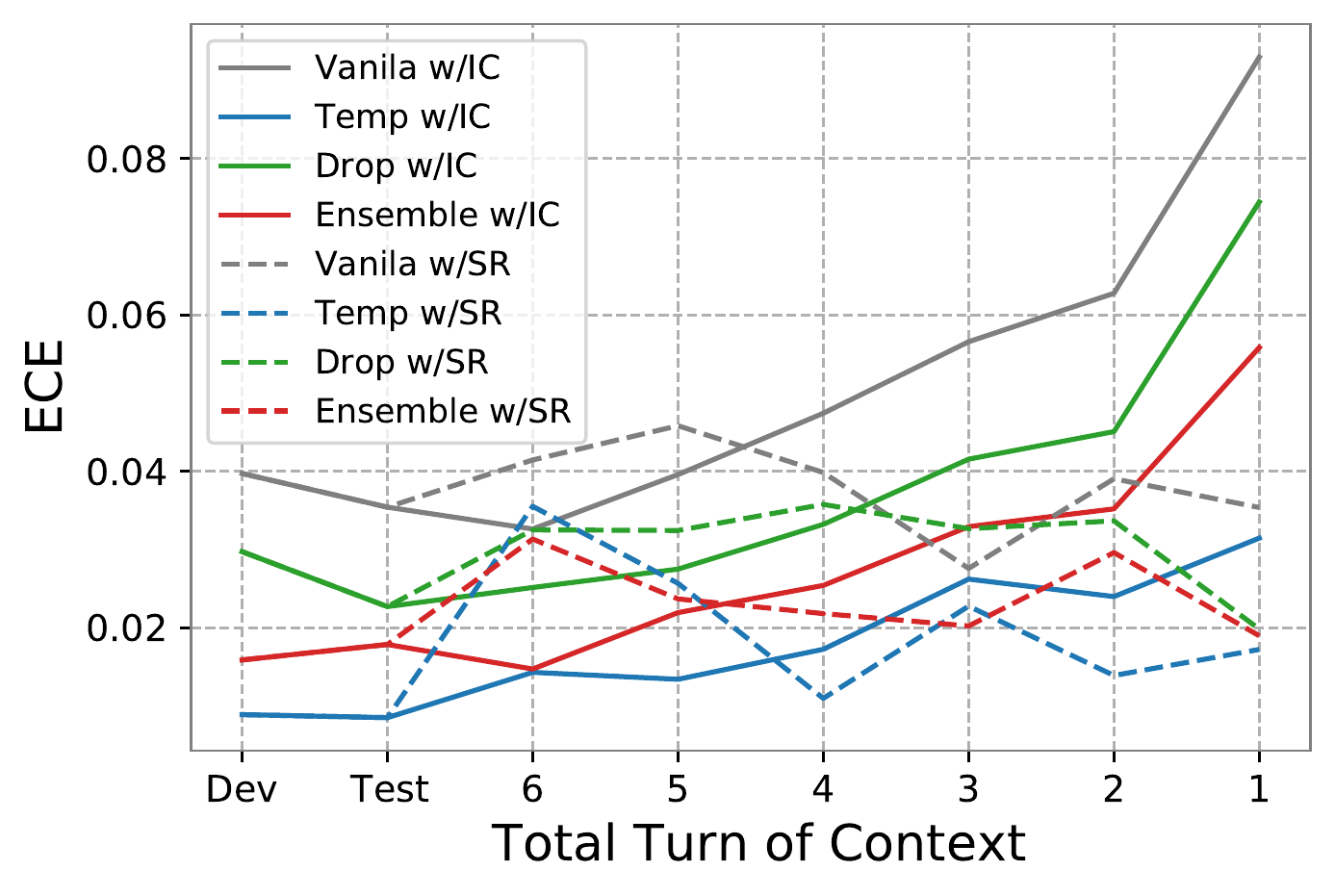}
        \caption[]%
        {{\small Deletion Ratio vs ECE}}    
        \label{fig:IC_ece}
    \end{subfigure}
    \caption[]
    {Results on the Accuracy (R@1/10), Brier score, and ECE of neural rankers under the distributional shifts with two methods and 2 comparisons, $UW$, $KW$ (top) and $IC$, $SR$ (bottom). Tables for numerical comparisons are provided in Appendix~\ref{appendixB} and~\ref{appendixC}.} 
    \label{fig:UW_IC}
\end{figure*}

\subsection{Calibration and accuracy of neural rankers under distributional shifts}
\label{exp_calibration}
We first verify the calibration and accuracy of neural rankers under the gradual distributional shifts of the source dialogue dataset. Figure~\ref{fig:UW_IC} shows the results for experiments that are intensifying the distributional shift by increasing the word replacement ratio of $UW$ and sentence deletion ratio of $IC$, respectively. Experimental details are provided in Appendix~\ref{appendixA}.

We can confirm that the accuracy decreases as the shift intensifies for both methods in Figures~\ref{fig:UW_acc} and~\ref{fig:IC_acc}\footnote{The vanilla model and the model with temperature scaling have the same accuracy, and have subtle differences in the Brier Score.}. However, as accuracy decreases, the Brier score and ECE gradually increase (see Figures~\ref{fig:UW_brier} and~\ref{fig:IC_brier}). Such results suggest that calibration for validation and training distributions does not guarantee calibration for distributional shifts. Overall, we show that the ensembles have higher robustness under distributional shifts, and that it is useful to find answers using collective predictions in dialogues. In addition, a post-hoc calibration method, the temperature scaling can be an efficient solution for maintaining calibration under distributional shifts (see Figures~\ref{fig:UW_ece} and~\ref{fig:IC_ece}).

We also evaluate the accuracy and calibration of neural rankers according to the importance of replacement words in the $UW$ (see Appendix~\ref{appendixA} and~\ref{appendixD}). The more important words are replaced in the dialogue context, the downgrade accuracy the rankers are. Significant observations are that the ensembles and the temperature scaling do not significantly decrease the ECE even if the target words become more important, indicating that both methods effectively calibrate neural rankers.

\subsection{Validity of distributional shift methods}
We analyze the validity of proposed distributional shift methods. A comparison group for the $UW$, Known Word ($KW$)\footnote{We assume the unknown word is a word that has appeared more than 5000 times in the training dataset.} is a method for replacing target words with known words that frequently occurred in the training dataset. We also set a Source ($SR$), which is a comparison group for the $IC$ to verify whether the calibration of neural rankers is affected by the number of turns in the dialogue context. Detailed explanations are provided in Appendix~\ref{appendixA}.

We confirm that $KW$ showed better performance than $UW$ as shown in Figures~\ref{fig:UW_acc},~\ref{fig:UW_brier}, and~\ref{fig:UW_ece}. The reason why $KW$ tends to be similar to $UW$ even though the target word has been replaced with the known words is that the meaning of the context has often been corrupted by incorrect word replacements. Experiments on $SR$ show that accuracy and calibration were maintained even when the source dataset became shorter, indicating that the calibration of the neural rankers was irrelevant to the length of the dialogue as shown in Figures~\ref{fig:IC_acc},~\ref{fig:IC_brier}, and~\ref{fig:IC_ece}. Consequently, methods for deleting sentences or replacing words could shift data distribution toward interfering with the calibration.
\section{Conclusion}
\label{conclusion}
We presented evaluations of the accuracy and calibration of various uncertainty estimation methods under the gradual distributional shifts of the DailyDialog dataset. Despite performance degradation under distributional shift, the ensembles seem to achieve the best across overall metrics and are more robust to distributional shift. Besides, the post-hoc calibration method, temperature scaling could be an efficient solution for robustness under the shift. We proved that replacing essential words or deleting sentences are valid methods to make the distributional shifts in the dialogue dataset. The dialogue systems should be evaluated under the extensive dataset shift, and we hope our proposed methods useful for evaluating the systems.

\bibliography{emnlp2021}

\begin{thebibliography}{25}
\expandafter\ifx\csname natexlab\endcsname\relax\def\natexlab#1{#1}\fi

\bibitem[{Amodei et~al.(2016)Amodei, Olah, Steinhardt, Christiano, Schulman,
  and Man{\'e}}]{amodei2016concrete}
Dario Amodei, Chris Olah, Jacob Steinhardt, Paul Christiano, John Schulman, and
  Dan Man{\'e}. 2016.
\newblock Concrete problems in ai safety.
\newblock \emph{arXiv preprint arXiv:1606.06565}.

\bibitem[{Brier(1950)}]{brier1950verification}
Glenn~W Brier. 1950.
\newblock Verification of forecasts expressed in terms of probability.
\newblock \emph{Monthly weather review}, 78(1):1--3.

\bibitem[{Br{\"o}cker(2009)}]{brocker2009reliability}
Jochen Br{\"o}cker. 2009.
\newblock Reliability, sufficiency, and the decomposition of proper scores.
\newblock \emph{Quarterly Journal of the Royal Meteorological Society: A
  journal of the atmospheric sciences, applied meteorology and physical
  oceanography}, 135(643):1512--1519.

\bibitem[{Devlin et~al.(2019)Devlin, Chang, Lee, and
  Toutanova}]{devlin-etal-2019-bert}
Jacob Devlin, Ming-Wei Chang, Kenton Lee, and Kristina Toutanova. 2019.
\newblock \href {https://doi.org/10.18653/v1/N19-1423} {{BERT}: Pre-training of
  deep bidirectional transformers for language understanding}.
\newblock In \emph{Proceedings of the 2019 Conference of the North {A}merican
  Chapter of the Association for Computational Linguistics: Human Language
  Technologies, Volume 1 (Long and Short Papers)}, pages 4171--4186,
  Minneapolis, Minnesota. Association for Computational Linguistics.

\bibitem[{Fellbaum(1998)}]{wordnet}
Christiane Fellbaum. 1998.
\newblock \emph{WordNet: An Electronic Lexical Database}.
\newblock Bradford Books.

\bibitem[{Feng et~al.(2020)Feng, Mehri, Eskenazi, and
  Zhao}]{feng-etal-2020-none}
Yulan Feng, Shikib Mehri, Maxine Eskenazi, and Tiancheng Zhao. 2020.
\newblock \href {https://doi.org/10.18653/v1/2020.acl-main.182} {{``}none of
  the above{''}: Measure uncertainty in dialog response retrieval}.
\newblock In \emph{Proceedings of the 58th Annual Meeting of the Association
  for Computational Linguistics}, pages 2013--2020, Online. Association for
  Computational Linguistics.

\bibitem[{Gal and Ghahramani(2016)}]{gal2016dropout}
Yarin Gal and Zoubin Ghahramani. 2016.
\newblock Dropout as a bayesian approximation: Representing model uncertainty
  in deep learning.
\newblock In \emph{international conference on machine learning}, pages
  1050--1059. PMLR.

\bibitem[{Gu et~al.(2020)Gu, Ling, and Liu}]{gu2020utterance}
Jia-Chen Gu, Zhen-Hua Ling, and Quan Liu. 2020.
\newblock Utterance-to-utterance interactive matching network for multi-turn
  response selection in retrieval-based chatbots.
\newblock \emph{IEEE/ACM Transactions on Audio, Speech, and Language
  Processing}, 28:369--379.

\bibitem[{Guo et~al.(2017)Guo, Pleiss, Sun, and
  Weinberger}]{guo2017calibration}
Chuan Guo, Geoff Pleiss, Yu~Sun, and Kilian~Q Weinberger. 2017.
\newblock On calibration of modern neural networks.
\newblock In \emph{International Conference on Machine Learning}, pages
  1321--1330. PMLR.

\bibitem[{Hendrycks and Dietterich(2019)}]{hendrycks2019benchmarking}
Dan Hendrycks and Thomas~G. Dietterich. 2019.
\newblock \href {https://openreview.net/forum?id=HJz6tiCqYm} {Benchmarking
  neural network robustness to common corruptions and perturbations}.
\newblock In \emph{7th International Conference on Learning Representations,
  {ICLR} 2019, New Orleans, LA, USA, May 6-9, 2019}. OpenReview.net.

\bibitem[{Kingma and Ba(2015)}]{kingma2014adam}
Diederik~P. Kingma and Jimmy Ba. 2015.
\newblock \href {http://arxiv.org/abs/1412.6980} {Adam: {A} method for
  stochastic optimization}.
\newblock In \emph{3rd International Conference on Learning Representations,
  {ICLR} 2015, San Diego, CA, USA, May 7-9, 2015, Conference Track
  Proceedings}.

\bibitem[{Kontogiorgos et~al.(2019)Kontogiorgos, Pereira, and
  Gustafson}]{estimation_tod}
Dimosthenis Kontogiorgos, Andre Pereira, and Joakim Gustafson. 2019.
\newblock \href {https://doi.org/10.1145/3340555.3353722} {Estimating
  uncertainty in task-oriented dialogue}.
\newblock In \emph{2019 International Conference on Multimodal Interaction},
  ICMI '19, page 414–418, New York, NY, USA. Association for Computing
  Machinery.

\bibitem[{Lakshminarayanan et~al.(2017)Lakshminarayanan, Pritzel, and
  Blundell}]{lakshminarayanan2016simple}
Balaji Lakshminarayanan, Alexander Pritzel, and Charles Blundell. 2017.
\newblock \href
  {https://proceedings.neurips.cc/paper/2017/hash/9ef2ed4b7fd2c810847ffa5fa85bce38-Abstract.html}
  {Simple and scalable predictive uncertainty estimation using deep ensembles}.
\newblock In \emph{Advances in Neural Information Processing Systems 30: Annual
  Conference on Neural Information Processing Systems 2017, December 4-9, 2017,
  Long Beach, CA, {USA}}, pages 6402--6413.

\bibitem[{Li et~al.(2017)Li, Su, Shen, Li, Cao, and Niu}]{li2017dailydialog}
Yanran Li, Hui Su, Xiaoyu Shen, Wenjie Li, Ziqiang Cao, and Shuzi Niu. 2017.
\newblock \href {https://www.aclweb.org/anthology/I17-1099} {{D}aily{D}ialog: A
  manually labelled multi-turn dialogue dataset}.
\newblock In \emph{Proceedings of the Eighth International Joint Conference on
  Natural Language Processing (Volume 1: Long Papers)}, pages 986--995, Taipei,
  Taiwan. Asian Federation of Natural Language Processing.

\bibitem[{Naeini et~al.(2015)Naeini, Cooper, and
  Hauskrecht}]{naeini2015obtaining}
Mahdi~Pakdaman Naeini, Gregory Cooper, and Milos Hauskrecht. 2015.
\newblock Obtaining well calibrated probabilities using bayesian binning.
\newblock In \emph{Proceedings of the AAAI Conference on Artificial
  Intelligence}, volume~29.

\bibitem[{Ovadia et~al.(2019)Ovadia, Fertig, Ren, Nado, Sculley, Nowozin,
  Dillon, Lakshminarayanan, and Snoek}]{NEURIPS2019_8558cb40}
Yaniv Ovadia, Emily Fertig, Jie Ren, Zachary Nado, D~Sculley, Sebastian
  Nowozin, Joshua Dillon, Balaji Lakshminarayanan, and Jasper Snoek. 2019.
\newblock \href
  {https://proceedings.neurips.cc/paper/2019/file/8558cb408c1d76621371888657d2eb1d-Paper.pdf}
  {{Can you trust your model$\backslash$textquotesingle s uncertainty?
  Evaluating predictive uncertainty under dataset shift}}.
\newblock In \emph{Advances in Neural Information Processing Systems},
  volume~32. Curran Associates, Inc.

\bibitem[{Penha and Hauff(2020)}]{penha2020curriculum}
Gustavo Penha and Claudia Hauff. 2020.
\newblock Curriculum learning strategies for ir.
\newblock In \emph{European Conference on Information Retrieval}, pages
  699--713. Springer.

\bibitem[{Penha and Hauff(2021)}]{penha2021calibration}
Gustavo Penha and Claudia Hauff. 2021.
\newblock On the calibration and uncertainty of neural learning to rank models
  for conversational search.
\newblock In \emph{Proceedings of the 16th Conference of the European Chapter
  of the Association for Computational Linguistics: Main Volume}, pages
  160--170.

\bibitem[{Petroni et~al.(2019)Petroni, Rockt{\"a}schel, Riedel, Lewis, Bakhtin,
  Wu, and Miller}]{petroni-etal-2019-language}
Fabio Petroni, Tim Rockt{\"a}schel, Sebastian Riedel, Patrick Lewis, Anton
  Bakhtin, Yuxiang Wu, and Alexander Miller. 2019.
\newblock \href {https://doi.org/10.18653/v1/D19-1250} {Language models as
  knowledge bases?}
\newblock In \emph{Proceedings of the 2019 Conference on Empirical Methods in
  Natural Language Processing and the 9th International Joint Conference on
  Natural Language Processing (EMNLP-IJCNLP)}, pages 2463--2473, Hong Kong,
  China. Association for Computational Linguistics.

\bibitem[{Qu et~al.(2018)Qu, Yang, Croft, Trippas, Zhang, and
  Qiu}]{qu2018analyzing}
Chen Qu, Liu Yang, W~Bruce Croft, Johanne~R Trippas, Yongfeng Zhang, and
  Minghui Qiu. 2018.
\newblock Analyzing and characterizing user intent in information-seeking
  conversations.
\newblock In \emph{The 41st international acm sigir conference on research \&
  development in information retrieval}, pages 989--992.

\bibitem[{van Niekerk et~al.(2020)van Niekerk, Heck, Geishauser, Lin, Lubis,
  Moresi, and Gasic}]{van-niekerk-etal-2020-knowing}
Carel van Niekerk, Michael Heck, Christian Geishauser, Hsien-chin Lin, Nurul
  Lubis, Marco Moresi, and Milica Gasic. 2020.
\newblock \href {https://doi.org/10.18653/v1/2020.findings-emnlp.277} {Knowing
  what you know: Calibrating dialogue belief state distributions via
  ensembles}.
\newblock In \emph{Findings of the Association for Computational Linguistics:
  EMNLP 2020}, pages 3096--3102, Online. Association for Computational
  Linguistics.

\bibitem[{Vig and Ramea(2019)}]{vig2019comparison}
Jesse Vig and Kalai Ramea. 2019.
\newblock Comparison of transfer-learning approaches for response selection in
  multi-turn conversations.

\bibitem[{Wolf et~al.(2020)Wolf, Debut, Sanh, Chaumond, Delangue, Moi, Cistac,
  Rault, Louf, Funtowicz, Davison, Shleifer, von Platen, Ma, Jernite, Plu, Xu,
  Scao, Gugger, Drame, Lhoest, and Rush}]{wolf2019huggingface}
Thomas Wolf, Lysandre Debut, Victor Sanh, Julien Chaumond, Clement Delangue,
  Anthony Moi, Pierric Cistac, Tim Rault, Rémi Louf, Morgan Funtowicz, Joe
  Davison, Sam Shleifer, Patrick von Platen, Clara Ma, Yacine Jernite, Julien
  Plu, Canwen Xu, Teven~Le Scao, Sylvain Gugger, Mariama Drame, Quentin Lhoest,
  and Alexander~M. Rush. 2020.
\newblock \href {https://www.aclweb.org/anthology/2020.emnlp-demos.6}
  {Transformers: State-of-the-art natural language processing}.
\newblock In \emph{Proceedings of the 2020 Conference on Empirical Methods in
  Natural Language Processing: System Demonstrations}, pages 38--45, Online.
  Association for Computational Linguistics.

\bibitem[{Wu et~al.(2017)Wu, Wu, Xing, Zhou, and Li}]{wu2017sequential}
Yu~Wu, Wei Wu, Chen Xing, Ming Zhou, and Zhoujun Li. 2017.
\newblock Sequential matching network: A new architecture for multi-turn
  response selection in retrieval-based chatbots.
\newblock In \emph{Proceedings of the 55th Annual Meeting of the Association
  for Computational Linguistics (Volume 1: Long Papers)}, pages 496--505.

\bibitem[{Yujian and Bo(2007)}]{yujian2007normalized}
Li~Yujian and Liu Bo. 2007.
\newblock A normalized levenshtein distance metric.
\newblock \emph{IEEE transactions on pattern analysis and machine
  intelligence}, 29(6):1091--1095.

\end{thebibliography}
\bibliographystyle{acl_natbib}

\appendix
\clearpage
\twocolumn
\section{Experiment Details}
\label{appendixA}
In this section, we present distributional shift methods and shifted data statistics following each method. We extend 1000 test instances of the DailyDialog dataset, constructing each turn of the dialogues as a response. The total number of extended test instances is 6639, and we use this dataset as the source dataset. We describe detailed experimental procedures as necessary. Our implementation is using a GeForce RT 3090 for training, and each epoch takes about 115 minutes. Our BERT-based ranker model has 110M parameters.

\subsection{UW: Word Replacement Ratio}
\label{ex_UW_WRR}

\begin{table}[h]
\centering
    {\small
    {\tabulinesep=0.6mm
    \begin{tabu}{c|cc|cc}
    & \multicolumn{2}{c|}{Unknown Word ($UW$)} & \multicolumn{2}{c}{Known Word ($KW$)}\\
    \hline
    ratio & \# of dataset & targets & \# of dataset & targets\\
    \hline
    0.05 & 5481 & 3.82 & 5481 & 3.82 \\
    0.10 & 6229 & 7.26 & 6229 & 7.26 \\
    0.15 & 6513 & 10.63 & 6512 & 10.64\\
    0.20 & 6609 & 14.20 & 6611 & 14.20\\
    0.25 & 6604 & 17.88 & 6618 & 17.85\\
    0.30 & 6550 & 21.57 & 6561 & 21.54\\
    0.35 & 6354 & 25.50 & 6456 & 25.38\\
    0.40 & 5831 & 30.02 & 6060 & 29.65\\
    0.45 & 4746 & 33.30 & 5137 & 33.41\\
    0.50 & 2787 & 32.99 & 3130 & 33.92\\
    \hline
    \end{tabu}
    }}
\caption{\label{table_UW_WRR} The total number of the dataset and the average number of replacement words, on the changes of the word replacement ratio.}
\end{table}

We follow the $UW$ method introduced in Section~\ref{method_UW}. We shift the source dataset by increasing the ratio of replacement words in the dialogue context.  The statistics in the shifted test dataset on the changes of the word replacement ratio are as shown in Table~\ref{table_UW_WRR}. We compare the accuracy and calibration performance of Vanilla, Temperature Scaling, Dropout, and Ensemble models on the 10 conditions.

\subsection{IC: Sentence Deletion Ratio}
\label{ex_IC}
We follow the $IC$ method introduced in Section~\ref{method_IC}. We shift the source dataset by increasing the sentence deletion ratio in the dialogue context. For the rigorous experiment, we filter and use only the source data with a context length of 6 turns. The total number of data is 2370, and the sentences are deleted one by one from the beginning, in order to construct the test dataset on each condition. We compare the accuracy and calibration performance of Vanilla, Temperature Scaling, Dropout, and Ensemble models on the 6 conditions.

\subsection{KW: Word Replacement Ratio}
\label{ex_KW}
In this experiment, we replace the target word with Known Word instead of Unknown Word for validation of the $UW$ method. We assume the Known Word is a word that has occurred more than 5000 times in the training dataset. The data statistics are shown in Table~\ref{table_UW_WRR}. Other experimental methods are the same as Section~\ref{ex_UW_WRR}.

\subsection{SR: Total Turn of Dialogue Context}
\label{ex_SR}
In this experiment, we use source dataset with a real context length of 6 turn to 1 turn for validation of the $IC$ method. Other experimental methods are the same as in Section~\ref{ex_IC}.

\subsection{UW: Word Importance}
\label{ex_UW_WI}

\begin{table}[h]
\centering
    {\small
    {\tabulinesep=0.6mm
    \begin{tabu}{c|cccc}
    \scriptsize{importance} & interval & \# of dataset & targets\\
    \hline
    1 & [0, 20) & 5925 & 6.58 \\
    2 & [20, 40) & 5925 & 6.98 \\
    3 & [40, 60) & 5925 & 6.98\\
    4 & [60, 80) & 5925 & 6.98\\
    5 & [80, 100) & 5925 & 7.39\\
    \hline
    \end{tabu}
    }}
\caption{\label{table_UW_WI} The attention interval, the total number of the dataset, and the average number of replacement words, on the changes of the word importance.}
\end{table}

In this experiment, we investigate the impact of the importance of replacement words on the accuracy and calibration of the model. We first fix the word replacement ratio at 20\%. We then sort the words(= token) in the dialogue context in order of each attention weight, and divide them into five buckets to rank their importance. For example, word importance 1 corresponds to the lower 20\% words in sorted words in the dialogue context. The data statistics are shown in Table~\ref{table_UW_WI}. We compare the accuracy and calibration performance of Vanilla, Temperature Scaling, Dropout, and Ensemble models on the 5 conditions (see Appendix~\ref{appendixD}).
\clearpage
\onecolumn
\section{Tables of Metrics (UW, IC)}
\label{appendixB}
The tables below report the Accuracy (R@1/10), Brier score, and ECE for each model and dataset are computed over $UW$ and $IC$ distributional shifts of the dataset.

\subsection{UW: Word Replacement Ratio}

\begin{table*}[h]
\centering
    {\small
    {\tabulinesep=0.8mm
    \begin{tabu}{l|ccc|ccc|ccc|ccc}
    Method & \multicolumn{3}{c|}{Vanilla} & \multicolumn{3}{c|}{Temp. Scaling} & \multicolumn{3}{c|}{Dropout} & \multicolumn{3}{c}{Ensemble}\\
    \hline
    Metrics & Acc & Brier & ECE & Acc & Brier & ECE & Acc & Brier & ECE & Acc & Brier & ECE \\
    \hhline{=|===|===|===|===}
    dev & 0.881 & 0.178 & 0.040 & 0.881 & 0.174 & 0.009 & 0.875 & 0.183 & 0.030 & 0.900 & 0.147 & 0.016\\
    test & 0.895 & 0.161 & 0.035 & 0.895 & 0.157 & 0.009 & 0.889 & 0.164 & 0.023 & 0.911 & 0.133 & 0.018\\
    \hline
    0.05 & 0.878 & 0.180 & 0.041 & 0.878 & 0.175 & 0.009 & 0.878 & 0.180 & 0.025 & 0.902 & 0.142 & 0.017\\
    0.10 & 0.859 & 0.206 & 0.041 & 0.859 & 0.202 & 0.010 & 0.855 & 0.208 & 0.025 & 0.886 & 0.166 & 0.012\\
    0.15 & 0.834 & 0.240 & 0.051 & 0.834 & 0.235 & 0.015 & 0.826 & 0.244 & 0.028 & 0.866 & 0.190 & 0.016\\
    0.20 & 0.790 & 0.305 & 0.072 & 0.790 & 0.295 & 0.018 & 0.781 & 0.308 & 0.046 & 0.835 & 0.238 & 0.024\\
    0.25 & 0.738 & 0.375 & 0.095 & 0.738 & 0.360 & 0.029 & 0.732 & 0.376 & 0.061 & 0.787 & 0.300 & 0.038\\
    0.30 & 0.688 & 0.450 & 0.115 & 0.688 & 0.429 & 0.039 & 0.671 & 0.456 & 0.089 & 0.740 & 0.362 & 0.051\\
    0.35 & 0.626 & 0.528 & 0.152 & 0.626 & 0.500 & 0.066 & 0.616 & 0.530 & 0.113 & 0.685 & 0.436 & 0.075\\
    0.40 & 0.544 & 0.647 & 0.198 & 0.544 & 0.605 & 0.111 & 0.528 & 0.648 & 0.169 & 0.605 & 0.542 & 0.112\\
    0.45 & 0.492 & 0.725 & 0.231 & 0.492 & 0.673 & 0.132 & 0.470 & 0.724 & 0.208 & 0.539 & 0.623 & 0.152\\
    0.50 & 0.441 & 0.796 & 0.259 & 0.441 & 0.737 & 0.156 & 0.424 & 0.798 & 0.233 & 0.487 & 0.699 & 0.183\\
    \hline
    \end{tabu}
    }}
    \caption{\label{table:UW_WRR} Numerical results in the Accuracy, Brier Score, and ECE of each model on the changes of the word replacement ratio in the dialogue context. The ratio of word replacement changes from 0.05 to 0.50 in 0.05 units.}
\end{table*}

\subsection{IC: Sentence Deletion Ratio}

\begin{table*}[h]
\centering
    {\small
    {\tabulinesep=0.8mm
    \begin{tabu}{l|ccc|ccc|ccc|ccc}
    Method & \multicolumn{3}{c|}{Vanilla} & \multicolumn{3}{c|}{Temp. Scaling} & \multicolumn{3}{c|}{Dropout} & \multicolumn{3}{c}{Ensemble}\\
    \hline
    Metrics & Acc & Brier & ECE & Acc & Brier & ECE & Acc & Brier & ECE & Acc & Brier & ECE \\
    \hhline{=|===|===|===|===}
    dev & 0.881 & 0.178 & 0.040 & 0.881 & 0.174 & 0.009 & 0.875 & 0.183 & 0.030 & 0.900 & 0.147 & 0.016\\
    test & 0.895 & 0.161 & 0.035 & 0.895 & 0.157 & 0.009 & 0.889 & 0.164 & 0.023 & 0.911 & 0.133 & 0.018\\
    \hline
    0/6 & 0.893 & 0.159 & 0.033 & 0.893 & 0.155 & 0.014 & 0.886 & 0.163 & 0.025 & 0.908 & 0.134 & 0.015\\
    1/6 & 0.885 & 0.170 & 0.040 & 0.885 & 0.166 & 0.013 & 0.879 & 0.175 & 0.028 & 0.897 & 0.147 & 0.022\\
    2/6 & 0.873 & 0.190 & 0.047 & 0.873 & 0.184 & 0.017 & 0.867 & 0.191 & 0.033 & 0.886 & 0.162 & 0.025\\
    3/6 & 0.852 & 0.219 & 0.057 & 0.852 & 0.212 & 0.026 & 0.846 & 0.221 & 0.042 & 0.873 & 0.186 & 0.033\\
    4/6 & 0.825 & 0.251 & 0.063 & 0.825 & 0.243 & 0.024 & 0.824 & 0.253 & 0.045 & 0.850 & 0.218 & 0.035\\
    5/6 & 0.742 & 0.368 & 0.093 & 0.742 & 0.351 & 0.031 & 0.738 & 0.366 & 0.074 & 0.762 & 0.323 & 0.056\\
    \hline
    \end{tabu}
    }}
    \caption{\label{table:IC_SDR} Numerical results in the Accuracy, Brier Score, and ECE of each model on the changes of the sentence deletion ratio in the dialogue context. The ratio of sentence deletion changes from 0 to 5/6 in 1/6 units.}
\end{table*}

\clearpage
\onecolumn
\section{Tables of Metrics (KW, SR)}
\label{appendixC}
The tables below report the Accuracy (R@1/10), Brier score, and ECE for each model and dataset are computed over $KW$ and $SR$ distributional shifts of the dataset.

\subsection{KW: Word Replacement Ratio}

\begin{table*}[h]
\centering
    {\small
    {\tabulinesep=0.8mm
    \begin{tabu}{l|ccc|ccc|ccc|ccc}
    Method & \multicolumn{3}{c|}{Vanilla} & \multicolumn{3}{c|}{Temp. Scaling} & \multicolumn{3}{c|}{Dropout} & \multicolumn{3}{c}{Ensemble}\\
    \hline
    Metrics & Acc & Brier & ECE & Acc & Brier & ECE & Acc & Brier & ECE & Acc & Brier & ECE \\
    \hhline{=|===|===|===|===}
    dev & 0.881 & 0.178 & 0.040 & 0.881 & 0.174 & 0.009 & 0.875 & 0.183 & 0.030 & 0.900 & 0.147 & 0.016\\
    test & 0.895 & 0.161 & 0.035 & 0.895 & 0.157 & 0.009 & 0.889 & 0.164 & 0.023 & 0.911 & 0.133 & 0.018\\
    \hline
    0.05 & 0.880 & 0.176 & 0.038 & 0.880 & 0.172 & 0.012 & 0.879 & 0.176 & 0.022 & 0.903 & 0.141 & 0.017\\
    0.10 & 0.868 & 0.196 & 0.037 & 0.868 & 0.193 & 0.013 & 0.859 & 0.198 & 0.024 & 0.895 & 0.154 & 0.011\\
    0.15 & 0.843 & 0.227 & 0.048 & 0.843 & 0.222 & 0.012 & 0.838 & 0.229 & 0.029 & 0.875 & 0.179 & 0.018\\
    0.20 & 0.814 & 0.274 & 0.060 & 0.814 & 0.268 & 0.023 & 0.810 & 0.275 & 0.036 & 0.853 & 0.214 & 0.021\\
    0.25 & 0.760 & 0.347 & 0.086 & 0.760 & 0.334 & 0.026 & 0.757 & 0.343 & 0.060 & 0.809 & 0.273 & 0.034\\
    0.30 & 0.720 & 0.401 & 0.100 & 0.720 & 0.385 & 0.033 & 0.714 & 0.401 & 0.068 & 0.770 & 0.320 & 0.044\\
    0.35 & 0.667 & 0.474 & 0.127 & 0.667 & 0.452 & 0.049 & 0.659 & 0.472 & 0.098 & 0.725 & 0.384 & 0.059\\
    0.40 & 0.603 & 0.569 & 0.165 & 0.603 & 0.536 & 0.078 & 0.594 & 0.561 & 0.133 & 0.662 & 0.471 & 0.091\\
    0.45 & 0.549 & 0.641 & 0.194 & 0.549 & 0.600 & 0.103 & 0.539 & 0.633 & 0.165 & 0.606 & 0.545 & 0.122\\
    0.50 & 0.505 & 0.704 & 0.213 & 0.505 & 0.657 & 0.114 & 0.491 & 0.688 & 0.190 & 0.558 & 0.602 & 0.140\\
    \hline
    \end{tabu}
    }}
    \caption{\label{table:KW_WRR} Numerical results in the Accuracy, Brier Score, and ECE of each model on the changes of the word replacement ratio in the dialogue context. The ratio of word replacement changes from 0.05 to 0.50 in 0.05 units.}
\end{table*}

\subsection{SR: Total Turn of Dialogue Context}

\begin{table*}[h]
\centering
    {\small
    {\tabulinesep=0.8mm
    \begin{tabu}{l|ccc|ccc|ccc|ccc}
    Method & \multicolumn{3}{c|}{Vanilla} & \multicolumn{3}{c|}{Temp. Scaling} & \multicolumn{3}{c|}{Dropout} & \multicolumn{3}{c}{Ensemble}\\
    \hline
    Metrics & Acc & Brier & ECE & Acc & Brier & ECE & Acc & Brier & ECE & Acc & Brier & ECE \\
    \hhline{=|===|===|===|===}
    dev & 0.881 & 0.178 & 0.040 & 0.881 & 0.174 & 0.009 & 0.875 & 0.183 & 0.030 & 0.900 & 0.147 & 0.016\\
    test & 0.895 & 0.161 & 0.035 & 0.895 & 0.157 & 0.009 & 0.889 & 0.164 & 0.023 & 0.911 & 0.133 & 0.018\\
    \hline
    6 & 0.908 & 0.141 & 0.041 & 0.908 & 0.139 & 0.036 & 0.904 & 0.141 & 0.033 & 0.919 & 0.118 & 0.031\\
    5 & 0.894 & 0.156 & 0.046 & 0.894 & 0.149 & 0.026 & 0.892 & 0.150 & 0.032 & 0.912 & 0.127 & 0.024\\
    4 & 0.884 & 0.171 & 0.040 & 0.884 & 0.169 & 0.011 & 0.873 & 0.180 & 0.036 & 0.906 & 0.135 & 0.022\\
    3 & 0.904 & 0.142 & 0.028 & 0.904 & 0.143 & 0.023 & 0.900 & 0.155 & 0.033 & 0.919 & 0.117 & 0.020\\
    2 & 0.887 & 0.170 & 0.039 & 0.887 & 0.166 & 0.014 & 0.874 & 0.179 & 0.034 & 0.895 & 0.149 & 0.030\\
    1 & 0.886 & 0.165 & 0.035 & 0.886 & 0.162 & 0.017 & 0.887 & 0.166 & 0.020 & 0.911 & 0.127 & 0.019\\
    \hline
    \end{tabu}
    }}
    \caption{\label{table:SR_SDR} Numerical results in the Accuracy, Brier Score, and ECE of each model on the changes of the length (turn) of the dialogue context. The length changes from 6 turn to 1 turn.}
\end{table*}

\clearpage
\onecolumn
\section{Figure and Table of Metrics (UW: Word Importance)}
\label{appendixD}
The figure and table below report the Accuracy (R@1/10), Brier score, and ECE for each model and dataset are computed over $UW$ distributional shifts on changes of word importance.

\subsection{Figure}
\begin{figure*}[h]
    \centering
    \begin{subfigure}[b]{0.32\textwidth}
        \centering
        \includegraphics[width=\textwidth]{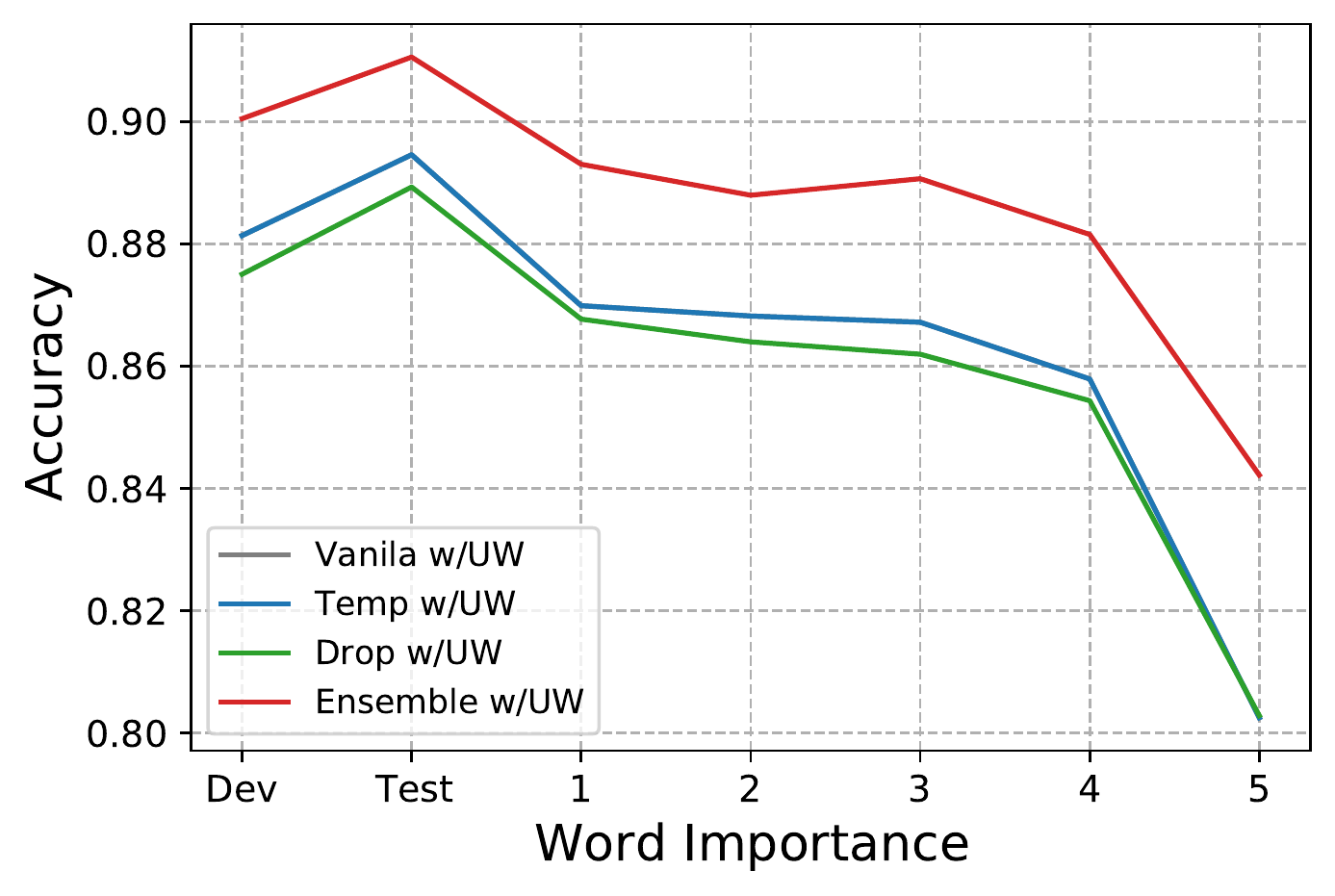}
        \caption[]%
        {{\small Word Importance vs Accuracy}}    
        \label{fig:UW_att_acc}
    \end{subfigure}
    \hfill
    \begin{subfigure}[b]{0.32\textwidth}  
        \centering 
        \includegraphics[width=\textwidth]{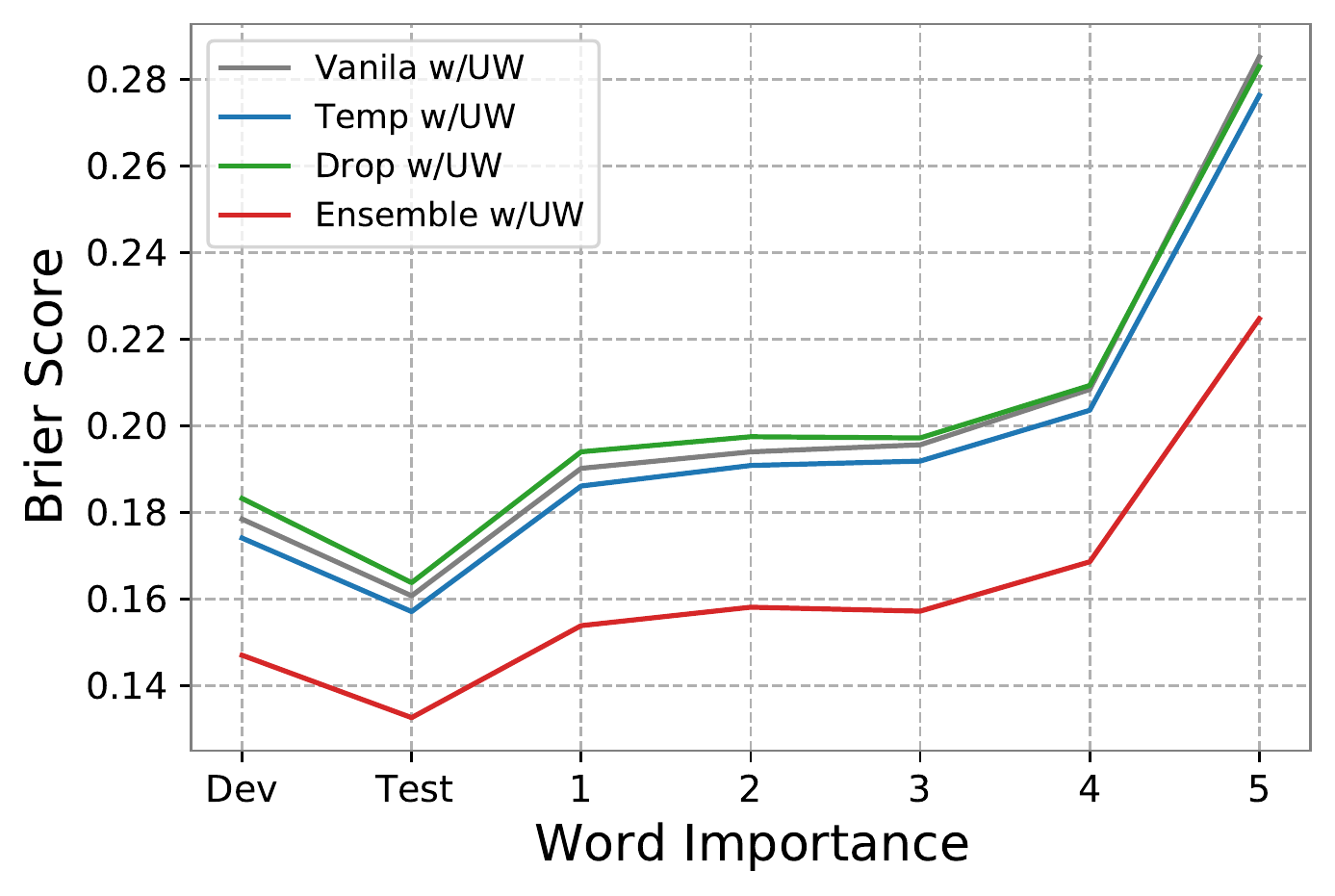}
        \caption[]%
        {{\small Word Importance vs Brier Score}}    
        \label{fig:UW_att_brier}
    \end{subfigure}
    \begin{subfigure}[b]{0.32\textwidth}
        \centering
        \includegraphics[width=\textwidth]{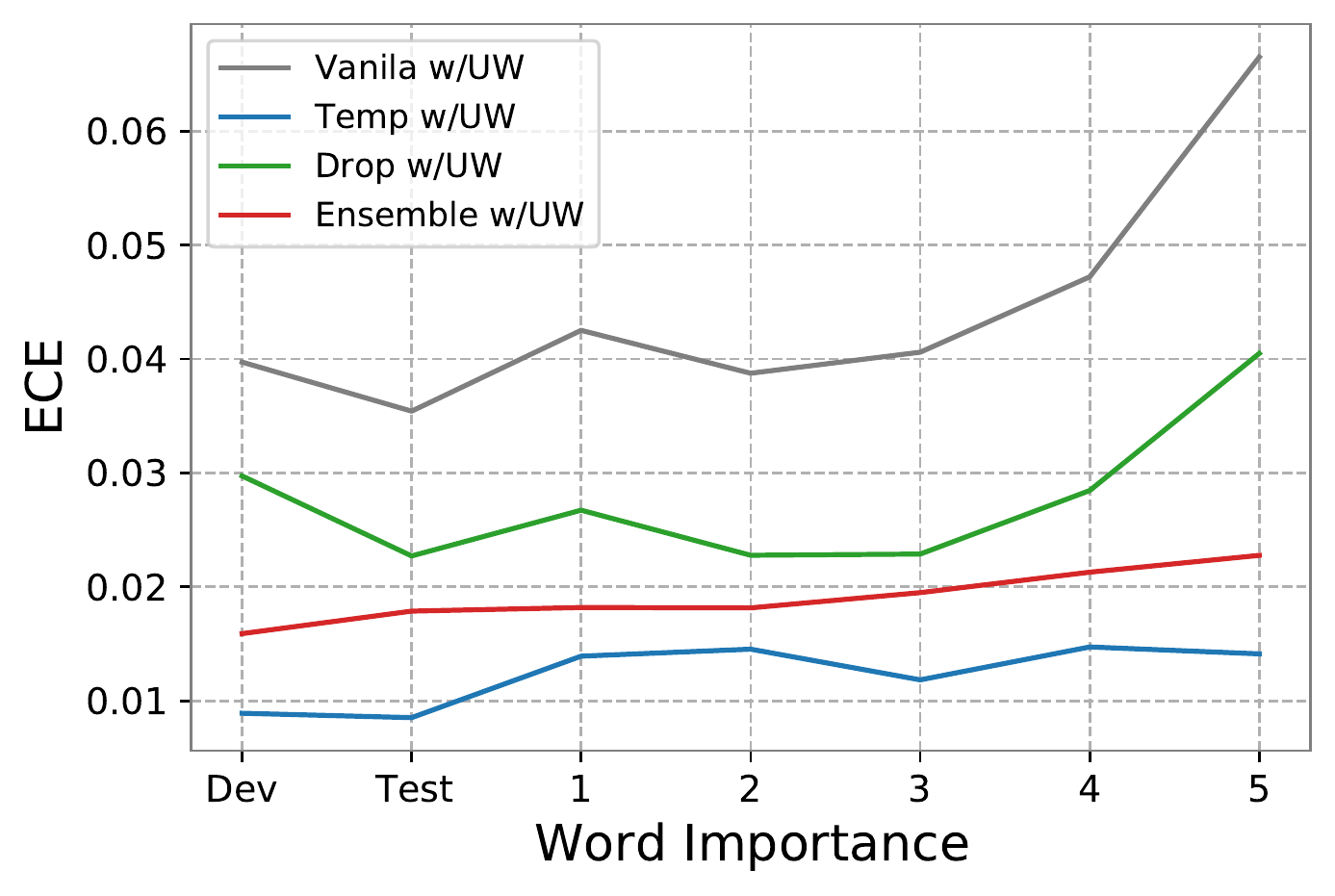}
        \caption[]%
        {{\small Word Importance vs ECE}}    
        \label{fig:UW_att_ece}
    \end{subfigure}
    \caption{\label{fig:UW_att} Results on the Accuracy, Brier score, and ECE of models under the distributional shifts with $UW$ strategy on the changes of the word importance.}
\end{figure*}

\subsection{Table of Metrics}
\begin{table*}[h]
\centering
    {\small
    {\tabulinesep=0.8mm
    \begin{tabu}{l|ccc|ccc|ccc|ccc}
    Method & \multicolumn{3}{c|}{Vanilla} & \multicolumn{3}{c|}{Temp. Scaling} & \multicolumn{3}{c|}{Dropout} & \multicolumn{3}{c}{Ensemble}\\
    \hline
    Metrics & Acc & Brier & ECE & Acc & Brier & ECE & Acc & Brier & ECE & Acc & Brier & ECE \\
    \hhline{=|===|===|===|===}
    dev & 0.881 & 0.178 & 0.040 & 0.881 & 0.174 & 0.009 & 0.875 & 0.183 & 0.030 & 0.900 & 0.147 & 0.016\\
    test & 0.895 & 0.161 & 0.035 & 0.895 & 0.157 & 0.009 & 0.889 & 0.164 & 0.023 & 0.911 & 0.133 & 0.018\\
    \hline
    1 & 0.870 & 0.190 & 0.043 & 0.870 & 0.186 & 0.014 & 0.868 & 0.194 & 0.027 & 0.893 & 0.154 & 0.018\\
    2 & 0.868 & 0.194 & 0.039 & 0.868 & 0.191 & 0.015 & 0.864 & 0.197 & 0.023 & 0.888 & 0.158 & 0.018\\
    3 & 0.867 & 0.196 & 0.041 & 0.867 & 0.192 & 0.012 & 0.862 & 0.197 & 0.023 & 0.891 & 0.157 & 0.019\\
    4 & 0.858 & 0.208 & 0.047 & 0.858 & 0.204 & 0.015 & 0.854 & 0.209 & 0.028 & 0.882 & 0.169 & 0.021\\
    5 & 0.803 & 0.285 & 0.067 & 0.803 & 0.276 & 0.014 & 0.803 & 0.283 & 0.040 & 0.842 & 0.225 & 0.023\\
    \hline
    \end{tabu}
    }}
    \caption{\label{table:UW_att_nu} Numerical results in the Accuracy, Brier Score, and ECE of each model on the changes of the word importance. The word importance changes from 1 to 5.}
\end{table*}

\end{document}